\documentclass[conference]{IEEEtran}
\IEEEoverridecommandlockouts
\usepackage[dvipsnames]{xcolor}

\usepackage{amsmath,graphicx,algorithmic,algorithm,booktabs,subcaption,multirow}

\usepackage{cite}
\usepackage{amsmath,amssymb,amsfonts}
\usepackage{algorithmic}
\usepackage{graphicx}
\usepackage{textcomp}
\usepackage{adjustbox}
\usepackage[absolute,overlay]{textpos}
\newlength{\TextAreaLeft}
\newlength{\TextAreaTop}
\newlength{\BottomMarginLen}
\newlength{\CopyrightBoxY}
\AtBeginDocument{%
  \setlength{\TextAreaLeft}{\dimexpr 1in+\hoffset+\oddsidemargin\relax}%
  \setlength{\TextAreaTop}{\dimexpr 1in+\voffset+\topmargin+\headheight+\headsep\relax}%
  \setlength{\BottomMarginLen}{\dimexpr \paperheight-\TextAreaTop-\textheight\relax}%
  \setlength{\CopyrightBoxY}{\dimexpr \textheight+\BottomMarginLen/3\relax}%
  \textblockorigin{\TextAreaLeft}{\TextAreaTop}%
}

\usepackage{hyperref}
\usepackage{xcolor}
\def\BibTeX{{\rm B\kern-.05em{\sc i\kern-.025em b}\kern-.08em
    T\kern-.1667em\lower.7ex\hbox{E}\kern-.125emX}}
\begin{document}

\title{Skeletonization-Based Adversarial Perturbations\\ on Large Vision Language Model's\\ Mathematical Text Recognition%

\thanks{This work was supported in part by the Japan Science and Technology Agency (JST) Support for Pioneering Research Initiated by the Next Generation (SPRING) under Grant JPMJSP2129; in part by the Japan Society for the Promotion of Science (JSPS) KAKENHI under Grant 25KJ2207 and 23K11174; and in part by the Ministry of Education, Culture, Sports, Science and Technology (MEXT), Japan, Promotion of Distinctive Joint Research Center Program under Grant JPMXP 0621467946.}
}

\author{
\IEEEauthorblockN{Masatomo Yoshida}
\IEEEauthorblockA{
Doshisha University\\
Kyoto, 610-0394  Japan \\
{\tt\small yoshida@vig.doshisha.ac.jp}
}
\\[.0em] 
\IEEEauthorblockN{Nicola Adami}
\IEEEauthorblockA{
University of Brescia\\
Brescia, 25134 Italy\\
{\tt\small nicola.adami@unibs.it}
}
\and
\IEEEauthorblockN{Haruto Namura}
\IEEEauthorblockA{
Independent Researcher\\
Tokyo, Japan \\
{\tt\small hrtnmr1@gmail.com}
}
\\[.0em]
\IEEEauthorblockN{Masahiro Okuda}
\IEEEauthorblockA{
Doshisha University\\
Kyoto, 610-0394  Japan \\
{\tt\small masokuda@mail.doshisha.ac.jp}
}
}

\maketitle
\begingroup\renewcommand\thefootnote{}

\endgroup
\begin{abstract}
This work explores the visual capabilities and limitations of foundation models by introducing a novel adversarial attack method utilizing skeletonization to reduce the search space effectively. Our approach specifically targets images containing text, particularly mathematical formula images, which are more challenging due to their LaTeX conversion and intricate structure. We conduct a detailed evaluation of both character and semantic changes between original and adversarially perturbed outputs to provide insights into the models’ visual interpretation and reasoning abilities. The effectiveness of our method is further demonstrated through its application to ChatGPT, which shows its practical implications in real-world scenarios.

\end{abstract}

\begin{textblock*}{\textwidth}(0pt, \CopyrightBoxY)
    \begin{adjustbox}{minipage=.98\textwidth,margin=.2cm,frame}
        \noindent
        \small
        \textbf{Accepted to ITC-CSCC 2025}\\
        \footnotesize{
        \copyright ~2025 IEEE. Personal use of this material is permitted. Permission from IEEE must be obtained for all other uses, in any current or future media, including reprinting/republishing this material for advertising or promotional purposes, creating new collective works, for resale or redistribution to servers or lists, or reuse of any copyrighted component of this work in other works. { }\href{https://doi.org/10.1109/ITC-CSCC66376.2025.11137646}{DOI: 10.1109/ITC-CSCC66376.2025.11137646}
        }
    \end{adjustbox}
\end{textblock*}

\section{Introduction}

The field of deep learning has significantly impacted image analysis, including text recognition tasks. Large Language Models (LLMs) such as ChatGPT \cite{achiam2023gpt} have expanded to include vision recognition, enabling the processing of images.

Adversarial attacks, which subtly manipulate input images to cause machine learning models to produce unintended outputs, expose vulnerabilities~\cite{Goodfellow2015, chen2017zoo, yoshida2022adversarial, namura2023effects}. While most research has focused on natural images, text recognition models, especially for mathematical expressions translated into LaTeX code, remain underexplored~\cite {wei2017adversarial}. The complexity of recognizing mathematical expressions and accurately converting them into LaTeX code makes these tasks significantly more challenging compared to simple text recognition.

\begin{figure}[!t]
    \centering
    \includegraphics[width=0.47\textwidth]{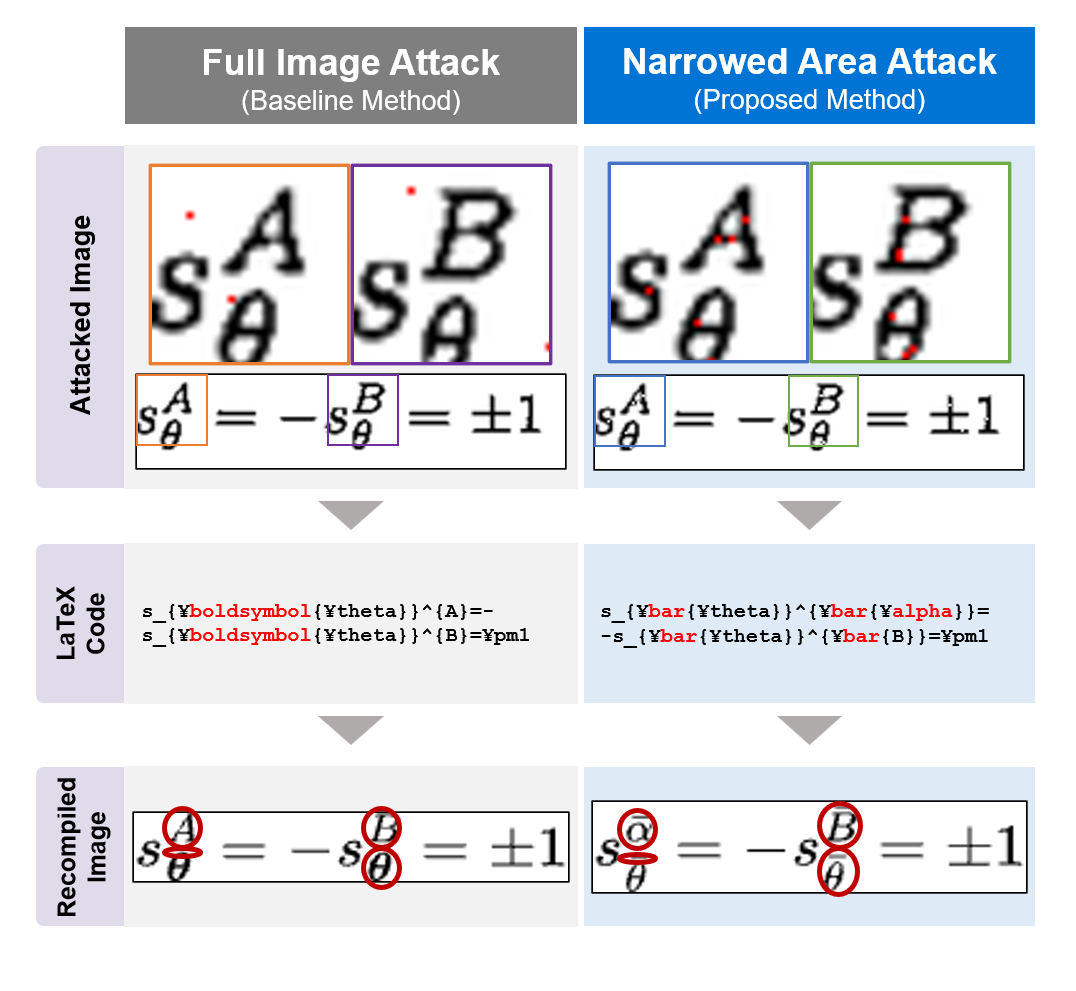}
    \caption{Overview of the proposed approach.} 
    \label{fig:exp_1}
    \vspace{-0.5em}
\end{figure}

\begin{figure*}[!t]
    \centering
    \includegraphics[width=0.9\textwidth]{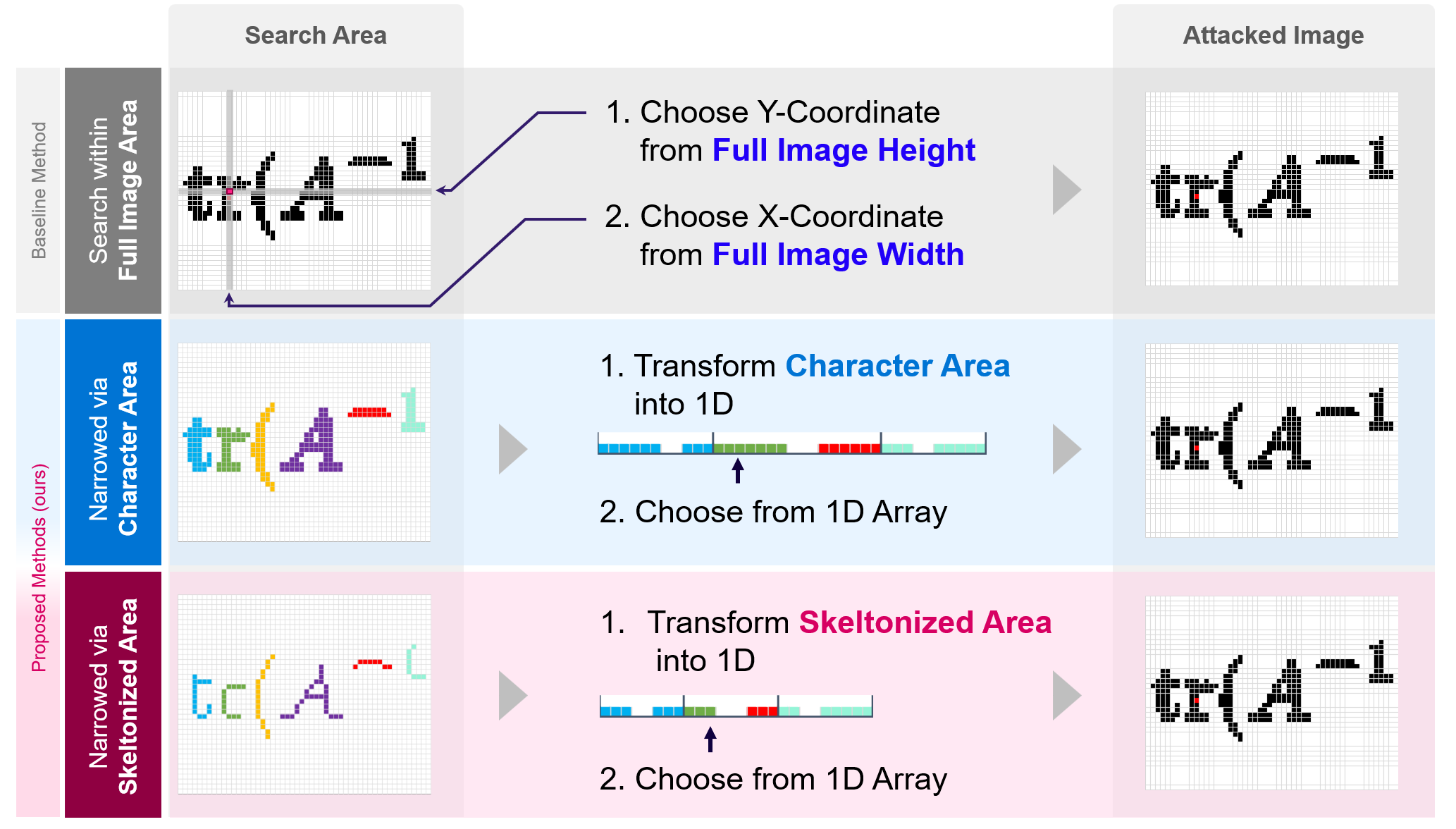}
    \caption{Comparison of the Process of the Target Pixel between Full Image Attack, Character Targeted Attack, and Attack on Skeletonized Area. The last two methods use the bounding box to form a 1D array.}
    \label{fig:narrow_comp}
\end{figure*}

Adversarial examples are crafted to exploit model weaknesses and improve robustness through adversarial training \cite{bai2021recent, Goodfellow2015}. This helps models withstand perturbations, enhancing their performance and security. However, the emergent visual capabilities of vision-capable foundation models reveal limitations that current benchmarks often fail to capture, necessitating innovative evaluation frameworks.

Our work addresses these gaps by proposing a novel adversarial attack method using skeletonization to reduce the search space. By focusing on text areas in mathematical formula images, our method is effective in black-box settings. We demonstrate its effectiveness through cosine similarity evaluation and validate it by transferring adversarial images to ChatGPT. This research also lead to enhance the robustness of foundation models in practical applications, such as preventing academic dishonesty in education, where tools like ChatGPT interpret mathematical content.

\section{Related Works and Background}

In the field of computer vision using deep learning models, adversarial examples are employed both for attacking models and for improving them through adversarial training~\cite{bai2021recent}. Foundational methods like the Fast Gradient Sign Method (FGSM)~\cite{Goodfellow2015} are classified as white-box attacks, which require access to model parameters and architecture. Conversely, query-based attacks~\cite{brendel2018decision, dong2019efficient} are categorized as black-box attacks, which operate without access to the internal workings of the model. Although more realistic in their application, black-box attacks typically entail higher computational costs.

Compared to general image classification tasks, adversarial examples for character recognition tasks pose unique challenges ~\cite{cahn2001mathematical,suzuki2003infty,deng2017image,mahdavi2019icdar}. This complexity is particularly evident in tasks involving the recognition of mathematical formula images, where the recognized image is subsequently converted into LaTeX code. This added layer of complexity surpasses that of simple character recognition tasks, introducing additional obstacles. Despite recent significant improvements in LaTeX character recognition (or LaTeX OCR), this area remains fraught with challenges. Models like Mathpix and pix2tex~\footnote{https://github.com/lukas-blecher/LaTeX-OCR}, which use vision transformers (ViT)~\cite{dosovitskiy2020image} and convolutional neural networks (CNN)~\cite{he2016deep}, have enhanced the accuracy of mathematical formula recognition. However, the robustness of these models against adversarial attacks remains under-explored. Given the complexity of mathematical notation and the necessity for precise spatial relationships, models in this field are particularly vulnerable to adversarial perturbations, underscoring the need for focused research.

Large Language Models (LLMs) like ChatGPT are classified as foundation models with vision recognition capabilities~\cite{achiam2023gpt, li2024multimodal}. These models offer new value to a wide range of users by integrating image recognition, but their limitations are not fully understood and require further grounding~\cite{kasneci2023chatgpt}. This research aims to contribute to the understanding of these limitations and promote better model usage by highlighting areas that need further investigation and improvement.

\section{Proposed Method}

Our proposed method, inspired by the One Pixel-Attack Method~\cite{su2019one}, introduces a novel approach to adversarial attacks on vision recognition models, specifically targeting mathematical formula recognition. %

\subsection{Skeletonization for Search Space Reduction}

The skeletonization process transforms input images into one-dimensional (1D) arrays, significantly reducing the search space for adversarial attacks and enhances the efficiency of the adversarial attack. We investigated the impact of skeletonization by performing experiments in three different scenarios, as shown in Figure \ref{fig:narrow_comp}. Key steps are as follows:

\begin{itemize}
    \item \textbf{Character Bounding Box Detection}: Initially, we detect the bounding boxes of all characters within the input image. This step ensures that we focus on the text areas, which are critical for OCR accuracy.
    \item \textbf{Skeletonization}: Next, we apply a skeletonization algorithm to each detected character. Skeletonization reduces the characters to their minimal form, typically one-pixel-wide lines, which represent the essential structure of the text.
    \item \textbf{1D Array Transformation}: The skeletonized character regions are then converted into a 1D array by concatenating the pixel values row by row. Finally, these arrays are concatenated from top to bottom and left to right, producing a unique 1D array representation for each image.
\end{itemize}

By focusing on the text areas and utilizing skeletonization, we exploit the inherent sparsity of text images. 

\subsection{Attack and Optimization}

The adversarial attack procedure involves generating initial adversarial images and iteratively refining them to minimize the similarity between the clean and adversarial outputs, using the pix2tex model as the LaTeX OCR Model. The attack process is outlined in Algorithm \ref{alg:pixel_attack}, and key steps are described below:

\begin{itemize}
\item \textbf{Initialization}: We generate an initial adversarial image by randomly perturbing the pixels within the skeletonized text areas.
\item As \textbf{loss function}, we use cosine similarity between the LaTeX code  of the clean and adversarial images to evaluate the effectiveness of the attack. Cosine similarity is calculated using Term Frequency-Inverse Document Frequency (TF-IDF)~\cite{manning2009introduction} vectors of the LaTeX sequences.
\item For \textbf{optimization}, we employ three methods: Covariance Matrix Adaptation Evolution Strategy (CMA-ES)~\cite{hansen2016cma,hansen2006cma}, Tree-structured Parzen Estimator (TPE)~\cite{bergstra2011algorithms}, and Random Search. Each method iteratively updates the pixel positions in the adversarial image, aiming to minimize the cosine similarity, thus increasing the disparity between the clean and adversarial outputs.
\end{itemize}

\subsection{Evaluation}

We evaluated the effectiveness of our adversarial attacks based on two criteria: \textbf{character change} using cosine similarity and \textbf{semantic change} through manual assessment.

\begin{itemize}
\item \textbf{Character Change}: We first assessed the syntactic differences between the LaTeX code outputs of the clean and adversarial images. Cosine similarity, calculated using Term Frequency-Inverse Document Frequency (TF-IDF) vectors, was used as the loss function during optimization. 
\item \textbf{Semantic Change}: To obtain a more precise evaluation, we manually assessed whether the changes in the LaTeX code altered the actual meaning of the mathematical expressions. 
\end{itemize}

To validate the effectiveness and transferability of our optimized adversarial attacks, we input the adversarially perturbed images into ChatGPT, a foundation model with visual capabilities. We then assessed the failure rates by comparing ChatGPT's outputs for the clean and adversarial images. %

\begin{algorithm}[t]
\caption{Procedure of the Proposed Adversarial Attack}
\label{alg:pixel_attack}
        \begin{algorithmic}[1]
            \REQUIRE $\text{Original Image }"x", \text{Adversarial Image } "x_{\text{adv}}", \newline \text{LaTeX OCR Model } "\text{OCR}",\newline \text{Optimization Method }"\text{Opt}"$, \text{TF-IDF Vectorizer} "V"
            \ENSURE $x_{\text{updated}}$
                    \STATE $x_{\text{adv}} \leftarrow \text{RandomlyPerturb}(x_{\text{adv}})$
            \STATE $y \leftarrow \text{OCR}(x)$
            \STATE $y_{\text{adv}} \leftarrow \text{OCR}(x_{\text{adv}})$
            \STATE $C_{\text{sim}} \leftarrow \frac{V(y) \cdot V(y_{\text{adv}})}{\|V(y)\| \|V(y_{\text{adv}})\|}{~~}\text{ // Calculate Cosine Similarity }$
            \FOR{$k \leftarrow 1$ \TO $N$}
                \STATE $\text{ // Minimize } C_{\text{sim}}$
                \STATE $x_{\text{adv}}' \leftarrow \text{Opt}(C_{\text{sim}}, \text{\# of attack pixels}, \text{search space} )$
                \STATE $y_{\text{adv}}' \leftarrow \text{OCR}(x_{\text{adv}}')$
                \STATE $C_{\text{sim}}' \leftarrow \frac{V(y) \cdot V(y_{\text{adv}}')}{\|V(y)\| \|V(y_{\text{adv}}')\|}$
                \IF{$C_{\text{sim}}' < C_{\text{sim}}$}
                    \STATE $x_{\text{adv}} \leftarrow x_{\text{adv}}'$
                    \STATE $C_{\text{sim}} \leftarrow C_{\text{sim}}'$
                \ENDIF
                
            \ENDFOR
            \STATE $x_{\text{updated}} \leftarrow x_{\text{adv}}$
        \end{algorithmic}
\end{algorithm}

\begin{table*}[t]
\centering
\caption{Comparison among Narrowing Methods}
\label{tab:res_narrow}
\scalebox{1.15}[1.15]{
\begin{tabular}{c|ccc|ccc|ccc|ccc} \toprule
Metrics               & \multicolumn{3}{c|}{Cosine Similarity $\downarrow$}         & \multicolumn{3}{c|}{Success Rate $\uparrow$}              & \multicolumn{3}{c|}{Accuracy $\downarrow$}                  & \multicolumn{3}{c}{PSNR} \\ \midrule
\# of Attacked Pixels & 15            & 20            & 25            & 15            & 20            & 25            & 15            & 20            & 25            & 15     & 20     & 25     \\ \midrule
Full Image            & 0.91          & 0.90          & 0.92          & 0.83          & 0.83          & 0.80          & 0.56          & 0.53          & 0.52          & inf    & 63.82  & 62.93  \\
Character Area        & \textbf{0.80} & \textbf{0.77} & \textbf{0.74} & 0.88          & \textbf{0.98} & \textbf{0.98} & 0.41          & 0.43          & 0.55          & 57.32  & 55.81  & 62.93  \\
Skeltonized Area      & 0.81          & 0.82          & 0.80          & \textbf{0.95} & \textbf{0.98} & \textbf{0.98} & \textbf{0.38} & \textbf{0.31} & \textbf{0.30} & 57.88  & 56.35  & 55.54 \\ \bottomrule
\end{tabular}
}
\vspace{0.5em}
\end{table*}

\begin{table}[]
\centering
\caption{Comparison among Optimization Methods}
\label{tab:res_opt}
\scalebox{1.15}[1.15]{
\begin{tabular}{c|ccc|c} \toprule
Metrics               & \multicolumn{3}{c|}{Cosine Similarity $\downarrow$}         & \multirow{2}{*}{\begin{tabular}[c]{@{}c@{}}Time $\downarrow$\\ (sec)\end{tabular}} \\ \cmidrule{1-4}
\# of Attacked Pixels & 15            & 20            & 25            &                                                                       \\ \midrule
CMA-ES                & 0.81          & 0.82          & 0.80          & 138.51                                                                \\
TPE                   & \textbf{0.79} & 0.80          & 0.78          & 147.71                                                                \\
Random Search         & \textbf{0.79} & \textbf{0.77} & \textbf{0.75} & \textbf{131.68}   \\ \bottomrule                                                   
\end{tabular}
}
\vspace{0.5em}
\end{table}

\section{Experiment}

\subsection{Dataset Creation}

To evaluate our adversarial attack method, we developed a specialized dataset comprising 40 digital images of mathematical equations. These images were resized to heights 50 pixels to ensure consistent evaluation across test cases.

To avoid potential data leakage from existing datasets, particularly considering that models like ChatGPT might have used datasets such as im2latex-100k~\cite{deng2017image} for training, we created an entirely new dataset. This approach prevents any influence from previously seen data and maintains the integrity of our results.

Considering each image in our dataset can have multiple valid LaTeX representations, We finally accounted for several hundred potential variations, which ensures the dataset is comprehensive and robust enough to effectively evaluate our adversarial attacks.

\vspace{0.5em}
\subsection{Narrowing Methods Comparison}

In this experiment, we compared different narrowing methods for the attack areas. As a baseline method, we used the \textbf{full image area}. We also employed image processing techniques, such as extracting the \textbf{character area}, and applying \textbf{skeletonization}. By performing skeletonization, the text rea is reduced to one-pixel-wide lines, making it easier to transform into a 1D array.

The comparison results are shown in Table \ref{tab:res_narrow}. The Success Rate represents the percentage of images where the Cosine Similarity dropped below 1. The Accuracy is the ratio of the number of identical characters to the total number of characters. PSNR indicates the difference from the peak value before the attack. From the results in Table \ref{tab:res_narrow}, it was observed that the more the search space is narrowed, the more effective the attack becomes. This suggests that each pixel within the narrowed search space effectively influences the interpretation of the Vision-capabled Model.

\vspace{0.5em}
\subsection{Optimization Methods Comparison}

Next, we compared the Optimization Methods used together with narrowing the search space. The optimization methods employed were Covariance Matrix Adaptation Evolution Strategy (CMA-ES), Tree-structured Parzen Estimator (TPE), and Random Search. CMA-ES is a stochastic evolutionary strategy designed for multivariate nonlinear optimization problems. It adapts the sample distribution in the search space to converge to the optimal solution by optimizing the search direction and step size using a covariance matrix, making it robust for complex objective functions and global optimization problems. TPE is a Bayesian optimization technique effective for black-box function optimization, particularly in hyperparameter tuning within machine learning. It conducts efficient searches with limited evaluations. Random Search is a simple and general optimization method that randomly selects sample points, widely used due to its simplicity but potentially inefficient in large or complex search spaces.

From Table \ref{tab:res_opt}, it is observed that Random Search consistently proved to be the most effective optimization method. This outperformance seems to be attributed by its ability to explore a wider range and avoid getting trapped in local optima, compared to the other optimization methods.

\vspace{0.5em}
\subsection{Evaluation on Actual Service}

As the final experiment, we evaluated the Semantic Change by inputting the images obtained from the our optimized attacks on the pix2tex model into ChatGPT (GPT-4 Model on the web). We evaluated the LaTeX code recognized by ChatGPT from both original and attacked images. In Table \ref{res:actual-svc}, the Upper row shows the results of the LaTeX code recognized by ChatGPT from the original image, and the Lower row shows the one recognized from the attacked image. Our proposed method demonstrates significant effects in both Character change and Semantic change, indicating that black-box transfer attacks of our approach on real-world Foundation Model with Vision capabilities are effective.

\begin{table}[t]
\caption{Rates of Semantic Change and Averaged Character Change in Images when Processed through ChatGPT}
\label{res:actual-svc}
\centering
\scalebox{1.15}[1.15]{
\begin{tabular}{c|cc} \toprule
\multicolumn{1}{c|}{} & \begin{tabular}[c]{@{}c@{}}Character Chg. \\      (Cosine Sim. $\downarrow$ )\end{tabular} & \begin{tabular}[c]{@{}c@{}}Rate of    \\Semantic Chg. $\uparrow$\end{tabular} \\ \midrule
Original Image & 0.99 & 0.28 \\
Attacked Image & \textbf{0.92} & \textbf{0.70} \\ \bottomrule
\end{tabular}
}
\vspace{0.5em}
\end{table}

\vspace{0.5em}
\section{Conclusion}

This study presented a novel adversarial attack method targeting visual foundation models, specifically those capable of recognizing mathematical equations. By combining skeletonization with optimization techniques, our approach effectively reduced the search space and improved attack efficiency. The optimized attacks demonstrated significant impacts in both syntactic changes (character change) and semantic changes.

Our experimental results showed that narrowing the search space using skeletonization enhanced the effectiveness of adversarial attacks, as shown by lower cosine similarity scores and higher success rates. Additionally, the transferability of these attacks was validated by inputting the adversarial images into actual service (ChatGPT). The results highlighted substantial semantic changes, which indicates that our method showed its effectiveness in the real-world services operating in a black-box setting.

Based on our results, which revealed the limitations and biases of foundation models, it is expected that these insights can help improve the fairness and robustness of future models. Future work will focus on developing more advanced attack strategies, such as score-based query attacks, to further enhance the effectiveness of adversarial perturbations.

{\fontsize{12}{14}\selectfont
{
    \fontsize{12}{14}\selectfont %
    \bibliographystyle{IEEEbib}
    \bibliography{strings,refs}
}
}

\end{document}